\documentclass[sigconf]{acmart}
\renewcommand\footnotetextcopyrightpermission[1]{}

\def\eg{{\em e.g.}}
\def\etal{{\em et al.}}
\newcommand{\figref}[1]{Fig. \ref{#1}}
\newcommand{\tabref}[1]{Tab. \ref{#1}}

\newcommand{\secref}[1]{Sect. \ref{#1}}

\newcommand{\mc}[1]{\mathcal{#1}}
\newcommand{\mb}[1]{\mathbb{#1}}

\usepackage{ragged2e}
\usepackage{booktabs}
\usepackage{color}
\usepackage{multirow}
\usepackage{bm}
\usepackage{bbm}
\usepackage[misc]{ifsym}
\usepackage{multirow}
\usepackage{float}

\usepackage{amssymb}

\AtBeginDocument{%
  \providecommand\BibTeX{{%
    \normalfont B\kern-0.5em{\scshape i\kern-0.25em b}\kern-0.8em\TeX}}}

\copyrightyear{2021}
\acmYear{2021}

\setcopyright{acmcopyright}
\acmConference[MM '21]{Proceedings of the 29th ACM International Conference on Multimedia}{October 20--24, 2021}{Virtual Event, China}
\acmBooktitle{Proceedings of the 29th ACM International Conference on Multimedia (MM'21), October 20--24, Virtual Event, China}
\acmPrice{15.00}
\acmDOI{10.1145/3474085.3475283}
\acmISBN{978-1-4503-8651-7/21/10}

\begin{document}
\fancyhead{}
\author{Zhongxing Ma$^{1}$, Yifan Zhao$^{1}$, Jia Li$^{1,2*}$}
\thanks{$^{*}$Jia Li is the corresponding author (E-mail: \textsuperscript{}jiali@buaa.edu.cn).\\
Website: https://cvteam.net/}
\affiliation{%
	\institution{$^1$State Key Laboratory of Virtual Reality Technology and Systems, SCSE, Beihang University, Beijing, China}
	\institution{$^2$Peng Cheng Laboratory, Shenzhen, China}
	\country{}
	}

\title{Pose-guided Inter- and Intra-part Relational Transformer for Occluded Person Re-Identification}

\begin{abstract}
   Person Re-Identification (Re-Id) in occlusion scenarios is a challenging problem because a pedestrian can be partially occluded. The use of local information for feature extraction and matching is still necessary. Therefore, we propose a Pose-guided inter- and intra-part relational transformer (Pirt) for occluded person Re-Id, which builds part-aware long-term correlations by introducing transformer. In our framework, we firstly develop a pose-guided feature extraction module with regional grouping and mask construction for robust feature representations. The positions of a pedestrian in the image under surveillance scenarios are relatively fixed, hence we propose intra-part and inter-part relational transformer. The intra-part module creates local relations with mask-guided features, while the inter-part relationship builds correlations with transformers, to develop cross relationships between part nodes. With the collaborative learning inter- and intra-part relationships, experiments reveal that our proposed Pirt model achieves a new state of the art on the public occluded dataset, and further extensions on standard non-occluded person Re-Id datasets also reveal our comparable performances.
\end{abstract}

\maketitle

\begin{figure}[t]
	\begin{center}
		\includegraphics[width= \linewidth]{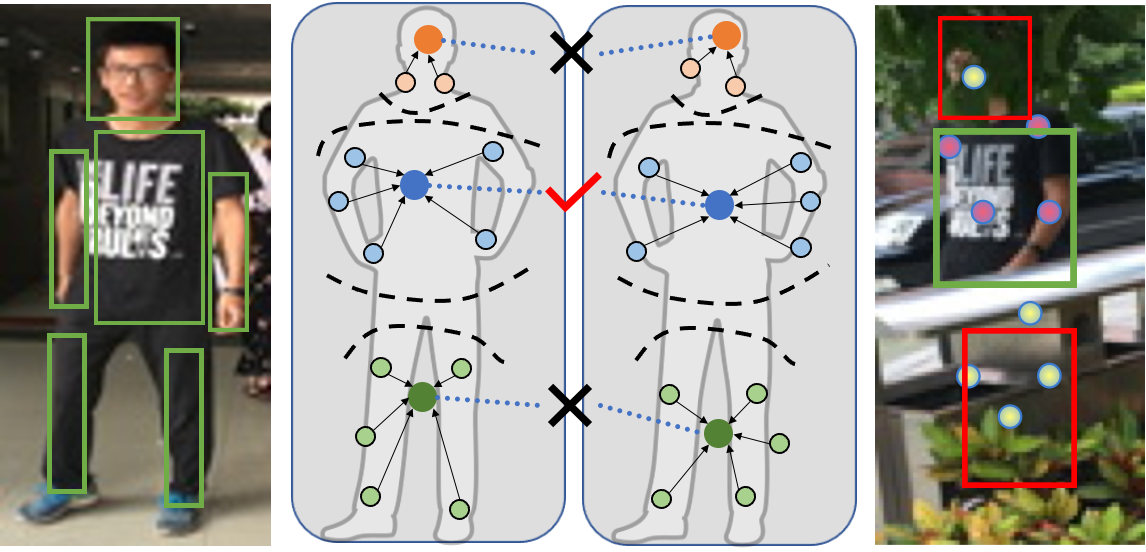}
	\end{center}
	\caption{The motivation of our proposed method. We estimate the keypoints of the pedestrian in image and extract local information with limited region. Then we aggregate information in more abstract part region according to the position of each keypoint. We capture the most representative information and match it with its corresponding semantic information to get the final retrieval results. }
    \label{fig:mot}
\end{figure}

\section{introduction}
Person Re-Identification, which intends to retrieve the most similar person image to the query person, has made great progress due to the wide application of deep learning \cite{reidsurvey-tpami}. However, handling cross-camera identification in complex occlusion scenes still faces great challenges. In the conventional person Re-Id task, the aligned person image usually contains the information of the holistic body \cite{market2015iccv, msmt172018cvpr}. Previous works \cite{pcb2018eccv, bdbnet2019iccv, raga2020cvpr, aanet2019cvpr} tend to build strong backbone features or add part correlations, thereby the robust global feature expressions of the person can be learned. By contrast, in the occlusion scene, there usually exists a large number of occluding objects in aligned person images \cite{occ2018icme}, which makes it difficult to produce reliable and discernible features.

By taking a deep investigation into the occluded person Re-Id task, the redundant occlusion does not only lead to the loss of target information but also introduces additional disturbing information. These occlusions of different characters such as color, size, shape, and structural position have affected the overall re-identification of the person. Most of the existing methods adopt the pose estimation \cite{horeid2020cvpr,pgfa2019iccv, fdgan2018nips}, the mask guidance \cite{gsfl2020eccv}, or the class activations \cite{dsr2018cvpr, fpr2019iccv, vpm2019cvpr, pa2017iccv, ado2018iccv, pb2018eccv} as auxiliary guidance to amplify the discriminative regions with higher confidence. Mask-guided information \cite{fpr2019iccv} helps discover the foreground region but loses the internal relation of keypoints which is provided by the pose estimation like \cite{horeid2020cvpr}.

To solve this problem, pose-guided person Re-Id models \cite{pgfa2019iccv, horeid2020cvpr, fdgan2018nips} adopt an auxiliary keypoint detector to detect visible body parts, such as heads, hands, and legs, which build the strong semantic representation of the object. In the occluded scenarios, the overlap of other agnostic objects usually leads to the disturbance of the right location of body parts. Hence directly matching these regions would produce severe errors in the retrieval phase. To this end, most of the existing methods \cite{pgfa2019iccv, horeid2020cvpr, fdgan2018nips} adopt local feature matching, such as feature alignment and graph matching.

Wang ~\etal~\cite{horeid2020cvpr} propose a Graph Convolutional Network (GCN) based method to generate part features using pose guided keypoints and they apply graph matching to align these part features. However, some keypoints extracted from the image cover the background and occlusion or they have limited area.  Miao ~\etal~\cite{pgfa2019iccv} propose a partial and pose-guided part feature alignment to select the useful information from the heavily occluded person. But the semantic information of different partial features is not always the same, and it neglects the relationship between these partial features.

Despite their performance differences, current occluded person Re-Id methods still show limitations in three aspects:
1) the pose estimation network generates unreliable localization, further feature matching with these parts is easily lead to misalignment;
2) the generation of local part regions usually neglects essential contextual information, leads to overfitting issues on visual patterns;
3) the structural relationship of keypoints structures is not deeply investigated, which makes it difficult to recognize some unreasonable matching results.

To solve the aforementioned problems, we propose a pose-guided inter- and intra-part relational transformer for the occluded person Re-Id task. For constructing reliable part features, we firstly expand the keypoint regions into larger masks and then merge the keypoint parts into several groups (3 groups for illustration in~\figref{fig:mot}). Thus further feature representation learning is conducted within each group to form a stable feature. Moreover, we aggregate the keypoint heatmaps to form a holistic object to enhance the foreground semantic regions for identification.

\begin{figure*}[t]
	\begin{center}
		\includegraphics[width=1\textwidth]{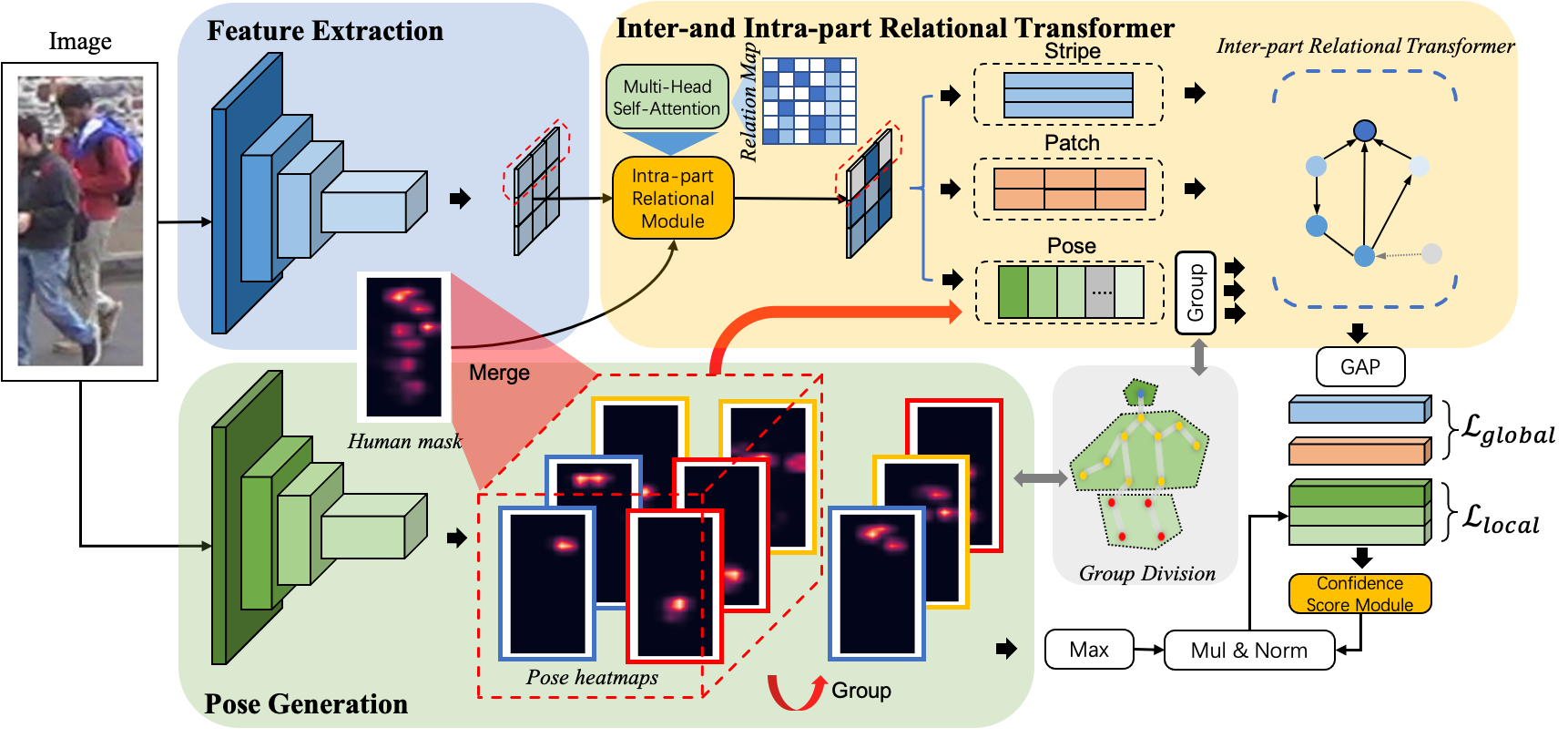}
	\end{center}
	\caption{The overall architecture of our proposed model, which consists of three essential parts: feature extraction from the image, keypoints generation from the image, inter- and intra-part relational transformer to aggregate the information between different visible or invisible regions.}
	\label{fig:arch}
\end{figure*}

After exerting pose-based information from an external pretrained model, we propose a joint part compositional model of three types of parts, encoding the contextual regions while strengthening the local features. We adopt the striped slice, patched grid, and pose-keypoint region as part representations of the image as shown in~\figref{fig:arch}. After introducing contextual information and building strong part representations, we propose our relational transformer to construct structural understanding. We adopt the successfully practiced transformer architecture in the field of natural language processing. With part parsing, each part feature is regarded as one graph node to adaptively learn the robust representation, which handles the missing semantics with the occlusion. In~\figref{fig:mot}, within each part group, every regularized keypoint region is aggregated into the holistic representation for this group. During the retrieval process, we search the nearest neighbor of the query image using global features and part-based local features.
With the collaborative constraints of local features and global features, our proposed Pirt model achieves a new state of the art on the public occluded person Re-Id dataset, Occluded-Duke dataset \cite{pgfa2019iccv} with the supervised setting. We also conduct detailed ablations to verify the effectiveness of the proposed pose guidance strategies and the inter- and intra-part relational architecture. Further extensions on standard non-occluded person Re-Id datasets also reveal the comparable generalization capability of our model.

Our contribution can be summarized as three-fold:
\vspace{-4pt}
\begin{enumerate}
    \item We propose a pose-guided inter- and intra-part relational transformer for occluded person re-identification, which builds long-term correlations by introducing transformer architectures. Experimental results also verify our proposed method reaches a new state of the art on public benchmarks.
    \item We make insightful improvements for pose-guided feature extractions, in which detected keypoints are expanded to construct the holistic object while forming part groups.
    \item We propose to learn the intra-part relationship with self-correlations and construct a multi-source inter-part relationship learning with relational transformers, providing a structural understanding of different part components.
\end{enumerate}
\vspace{10pt}

\section{Related Work}

\textbf{Person re-identification.} Great improvements have been made in supervised Person Re-Id \cite{reidsurvey-tpami}. One solution to handle the challenge is the part-based model, which was proposed according to the fixed position of one person in the image \cite{pcb2018eccv, spindle2017cvpr}. Sun~\etal~\cite{pcb2018eccv} proposed a uniform partitioning strategy to output the visual descriptor consisting of several horizontal part-level features, similar ideas can be founded in other fine-grained tasks \cite{zhao2021graph, he2019part}. On the other hand, to enhance the person feature representation capability of Convolutional Neural Network (CNN), several methods \cite{ianet2019cvpr, auto2019iccv, osnet2019iccv, bdbnet2019iccv} tried to extend modules to get better performance. Hou~\etal~\cite{ianet2019cvpr} proposed a spatial interaction-and-aggregation module to capture relationships between spatial features. Attention mechanism was used for designing CNNs to capture person information in images \cite{aanet2019cvpr, mixedo2019iccv, son2019iccv, raga2020cvpr, dndm2020eccv}. Zhang~\etal~\cite{raga2020cvpr} utilized the clustering-like information among spatial positions in the feature map and proposed two relation-aware global attention modules. In some mask-guided models \cite{mgcam2018cvpr, sgsnet2020cvpr}, external knowledge helps to capture the information of the foreground. In some pose-guided models \cite{dual2019iccv, posed2017iccv}, the positional and semantic information of the keypoints were used to produce local features and explore the connection between them.

\textbf{Occluded person re-identification.} Recognizing an occluded person is much difficult because of the confusing information and spatial feature misalignment \cite{horeid2020cvpr}. Most of the papers adopt the local feature matching method \cite{pa2017iccv, ado2018iccv, pb2018eccv, dsr2018cvpr, fpr2019iccv, pgfa2019iccv, horeid2020cvpr, pvpm2020cvpr, vpm2019cvpr, gsfl2020eccv, ig2020eccv}. He~\etal~\cite{dsr2018cvpr} proposed a method of spatial feature reconstruction that got robust local feature maps, then the author used the least square algorithm to solve the coefficient matrix to align corresponding spatial features. He~\etal~\cite{fpr2019iccv} proposed a method that was based on the inspiration of \cite{dsr2018cvpr}. The paper \cite{fpr2019iccv} proposed a method aiming to extract probability scores from the foreground probability network. The scores are used for the spatial reconstruction by assigning the body parts with confidence scores. Wang~\etal~\cite{horeid2020cvpr} proposed a keypoint semantic feature extraction method with pose guidance, then feature maps were passing messages using GNN. In addition, the method in the paper\cite{pgfa2019iccv} used the basic idea of the paper \cite{pcb2018eccv} with pose estimation methods to generate robust features, while \cite{gsfl2020eccv} used pose estimation and salience object detection simultaneously, which generated the semantic mask constrained by two keypoint detectors of the human body.

\textbf{Image retrieval.} The local feature matching used in the occluded person Re-Id is similar to the local feature matching used in image retrieval \cite{delf2017cvprw, delg2020eccv, superp2018iccvw, superg2020cvpr, dgm2018cvpr, cgm2019iccv}. Keypoints across images must comply some certain physical constraints. For example, one keypoint is associated with a certain keypoint in another image and some keypoints are mismatched due to occlusion. Sarlin~\etal~\cite{superg2020cvpr} proposed an attention-based graph neural network to jointly find corresponding keypoints and reject irrelevant keypoints. The paper \cite{superg2020cvpr} combined with traditional local feature detector and descriptor which extract sparse keypoints. Cao~\etal~\cite{delg2020eccv} introduced an auto-encoder-based dimensionality reduction technique for local features and combined the global feature with local features to get better performances. The local features were generated from CNN in the paper \cite{delg2020eccv}, which were a sort of high-level information extracted from the original image. Our method also draws some inspiration from the paper \cite{delg2020eccv}.

\textbf{Transformer.} The transformer was firstly proposed by \cite{attn2017nips} for machine translation. Similar and improved methods reached the state of the art in many natural language processing tasks. One of the representative works \cite{bert2018naacl} is a solid example to explain the strength of self-attention. With the development of the transformer, a lot of works \cite{imgtrans2018icml, esair2020cvpr} were looking at how to design the transformer for image recognition. \cite{esair2020cvpr} explored several self-attention modules which were invariant to permutation, then they proposed a patch-wise self-attention module with the capacity to uniquely identify specific locations.

\section{Method}
\subsection{Overview}

The difficulties of occluded person Re-Id lie in the incomplete information and the unstable location of the person in the image. In most cases, a pedestrian in the surveillance scenario has a standing or walking posture with his head, body, and legs well aligned from the top to bottom. An intricate occlusion environment would cause more negative effects, especially on locality. We use a pose estimation network $\mc{P}$ to extract keypoints of the person in the image $\mc{I} \in [0, 255]^{H\times W\times 3}$. The keypoints cover the corresponding positions of human body joints in the image and keypoints are represented by heatmaps $\mathbf{M} \in (0, 1)^{H_{m} \times W_{m} \times P}$, where $P$ is the number of keypoints. Besides, we use a Global Max Pooling (GMP) layer to expand the original limited coverage of keypoints to seize more essential contextual information. We extract the visible part information of the pedestrian through these heatmaps $\mathbf{M}$.

After the essential local part information is obtained, we merge all the heatmaps $\mathbf{M}$ into one human mask $\mathbf{C} \in (0, 1)^{H_f\times W_f}$. We combine human mask $\mathbf{C}$ together with the corresponding spatial feature map $\mathbf{F} \in \mb{R}^{H_f\times W_f\times C}$ generated by the backbone. $\mathbf{F}$ is passed into the Intra-part Relational Module (IRM) which is one of the components of our model. The feature maps $\mathbf{F}_{irm}$ generated by IRM are partitioned by different strategies into three-part feature sets as shown in \figref{fig:arch} and \figref{fig:ps}. For each feature in each set, we use a Global Average Pooling (GAP) layer to mix the information of the partitioned area. We denote $\mathbf{F}_{local} \in \mb{R}^{N\times C}$ as our local features and $N$ is the number of parts. In addition, all the keypoint parts are divided into three groups. We totally have three groups of keypoints that are represented by heatmaps: one group of head keypoints, one group of upper body keypoints, and the last one of lower body keypoints.

Finally, each type of local features $\mathbf{F}_{local}$ is sent into the Inter-part Relational Transform (IRT) to generate well representative features as the final embedding features. Embedding features partitioned by the $\mathbf{M}$ are received by Confidence Score Module (CSM). The outputs are incorporated with max score $\mathbf{S}$ from $\mathbf{M}$ to produce final embedding features.

\subsection{Pose-guided Feature Extraction}
\label{sec:1}

Person Re-Id is a branch of fine-grained task, therefore severe occlusion and multiple pedestrians in the image seriously impair the information of the target person. It is difficult for general models to learn the knowledge from the datasets and to reliably positioning the person in the image. To overcome the problems described above, pose information is critical to find keypoints that represent visible parts of the pedestrian.

We firstly use $\mc{P}$ to generate keypoints and their original heatmaps. For each heatmap, the value in each grid ranged from 0 to 1 represents the corresponding confidence score. Each initial pose heatmap only covers a limited region. Because the domain gap between the dataset trained for $\mc{P}$ and the dataset trained for Re-ID task is large. We use a GMP layer to expand the area and integrate more contextual information:
\begin{equation}
\label{eq:mask}
    \mathbf{M} = GMP(\mathbf{\hat{M}}),
\end{equation}
where $\mathbf{\hat{M}}$ is initial heatmaps generated by $\mc{P}$. After gathering all heatmaps $\mathbf{M}$ containing extra peripheral information, we choose the maximum confidence score $\mathbf{S}$ for all heatmaps on the identical grid to construct the human mask $\mathbf{C}$:
\begin{equation}
\label{eq:cob_mask}
    \mathbf{C}_{g} = max(\mathbf{M}_{g}),
\end{equation}
where $g$ represents the $g$th grid in original heatmaps. We use local features as complementary information to get better performances. For each local feature generated by each keypoint, we set a threshold $\tau$ to convert the original heatmaps $\mathbf{M}$ into 0-1 masks. Then we apply a GAP layer to generate features:
\begin{equation}
    \mathbf{F}_{pose} = GAP((\mathbf{M} \textgreater \tau) \cdot \mathbf{F}_{irm}),
\end{equation}
where $\mathbf{F}_{pose}$ is one type of $\mathbf{F}_{local}$ in \secref{sec:2}. We select the max confidence score in each grouped heatmaps $\mathbf{M}$ for subsequent fusion.

\subsection{Intra-part Relational Module}
With the guidance of pose information, learning methods for local part features from an image still have a large development space. Based on the mentioned traits of the image, we assume that each horizontal area probably contains part information of the pedestrian.
After we obtain the spatial features $\mathbf{F}$ generated by the backbone, we design a bottleneck-like style module called intra-part relational module. For each horizontal part of $\mathbf{F}$, our goal is to find the relationship between features in it. Our overall architecture of IRM is shown in \figref{fig:irm}. We split our architecture into three stages. At the first stage, we use a convolutional layer with $1\times1$ filters to encode the channel information. InstanceNorm layer, BatchNorm (BN) layer and the activation function ReLU to construct a block $\phi(\cdot)$, and for features $\mathbf{F}$:
\begin{figure}[t]
	\begin{center}
		\includegraphics[width=0.8\linewidth]{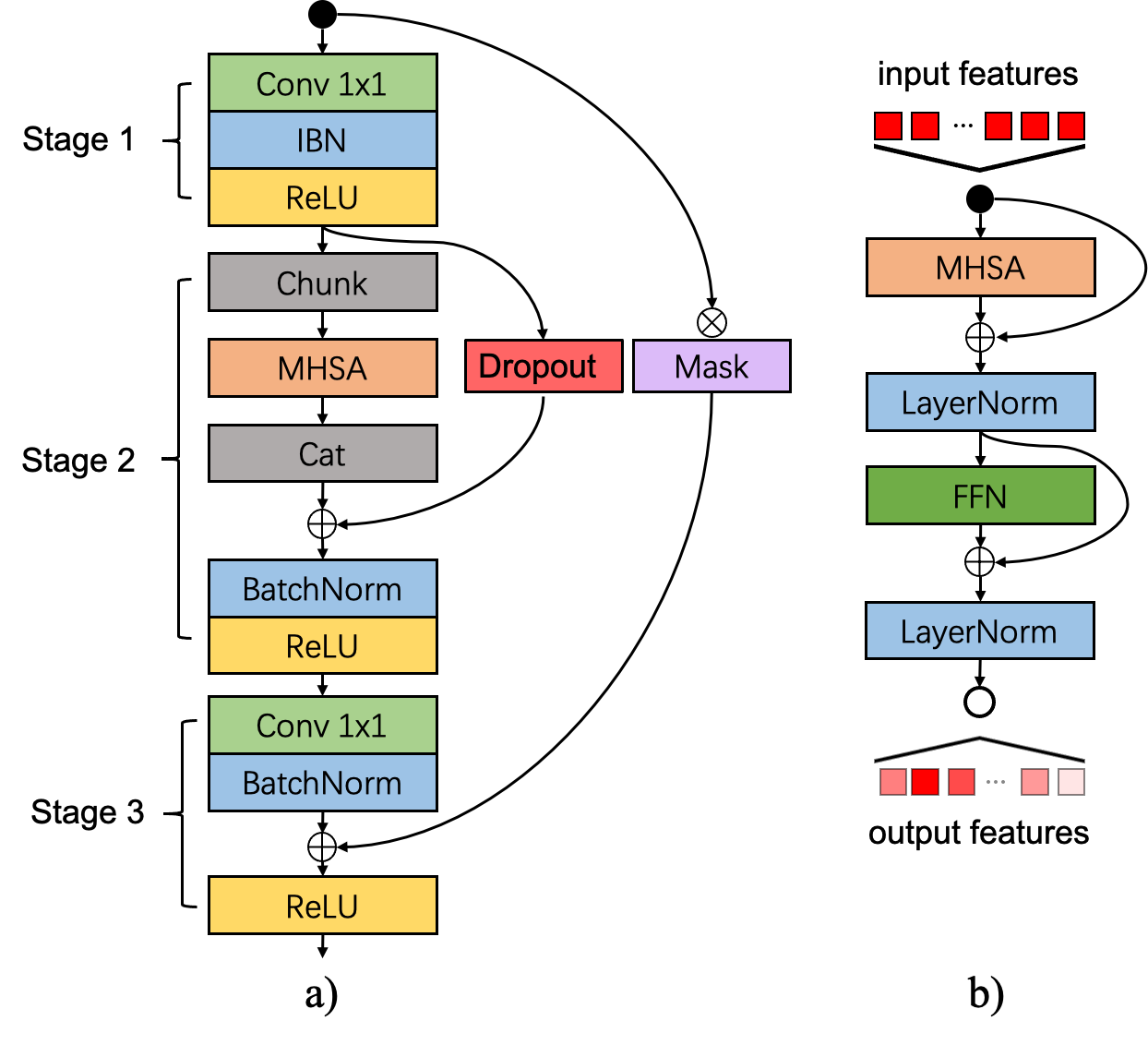}
	\end{center}
	\caption{Brief illustration of our intra-part relational module and transform unit. a) is our proposed intra-part relational module and b) is the general component of our inter-part relational transformer.}
	\label{fig:irm}
\end{figure}
\begin{equation}
    \mathbf{F}_{\phi} = \phi (\mathbf{F}),
\end{equation}
where $\mathbf{F}_{\phi} \in \mathbb{R}^{H_f\times W_f\times d}$ indicates the intermediate features generated by $\phi(\cdot)$, and $d$ is their dimension. To capture visible pedestrian information of each part, an attention mechanism called Multi-Head Self-Attention (MHSA) \cite{attn2017nips} is used to find the relation between features in the horizontal part. This module is used in a standard transformer block as well. We then horizontally divide $\mathbf{F}_{\phi}$ into striped features $\mathbf{\widetilde{F}}_{\phi} \in \mathbb{R}^{1\times W_f\times d}$ as shown in \figref{fig:irm}. The striped feature $\mathbf{\widetilde{F}}_{\phi}$ contains multiple individual features with dimension $d$.  We use MHSA to build relationships between these features, then the relational guidance is built to construct different attention scores for each feature $\mathbf{\widetilde{F}}_{\phi}$. We place all features into original position. To avoid the over-fitting problem that might be raised by MHSA, we adopt a Dropout layer to be a stochastic identical mapping:
\begin{equation}
    \mathbf{F}_{\phi}^{*}=concat(MHSA(chunk(\mathbf{F}_{\phi}))) + Dropout(\mathbf{F}_{\phi}).
\end{equation}
We apply BN and ReLU to generate final features $\mathbf{F}_{\phi}^{*}$ of stage 2. MHSA aggregates and captures important messages between features. The input consists of query features and key features of dimension ${d_k}$, and value features of dimension ${d_v}$. For self-attention, query features, key features, and value features are identical. We compute the relation matrix as same as \cite{attn2017nips}.

\begin{equation}
    \label{eq:mhsa}
	attn_i(\mathbf{X}) = softmax(\frac{\mathbf{X}{\mathbf{W}_i}^Q(\mathbf{X}{\mathbf{W}_i}^K)^T}{\sqrt{d_k}})\mathbf{X}{\mathbf{W}_i}^V,
\end{equation}

\begin{equation}
    \label{eq:mh}
	\mathbf{\overline{X}} = concat(attn_i(\mathbf{X}))\mathbf{W}^H \quad i \in 1,\ldots,h,
\end{equation}
where $\mathbf{X}$ indicates abstract input features, and $\mathbf{\overline{X}}$ indicates output features, and $\mathbf{W}^{Q, K, V}$ represent learnable weights of query, key, value features, and their projections. At last, we use $\mathbf{W}^{H}$ to enhance the features of one head. Applying multiple heads can get more relationships because of the diverse projections, and richer expressiveness of features through MHSA is then obtained. The next procedure of the final generated features $\mathbf{F}_{irm}$ is similar to function $\phi(\cdot)$. $\theta(\cdot)$ contains a convolutional layer and BN layer to restore original dimension as input features $\mathbf{F}$. Instead of applying the input features $\mathbf{F}$ as residual connection, we combine these features with human mask $\mathbf{C}$ as an auxiliary attention complement. Finally, $\mathbf{F}_{irm}$ is produced by the following equation:
\begin{equation}
    \mathbf{F}_{irm} =  ReLU (\theta (\mathbf{F}_{\phi}^{*}) + \mathbf{C} \cdot \mathbf{F}).
\end{equation}

\subsection{Inter-part Relational Transformer}
\label{sec:2}
In complex and extreme occlusion scenes, \eg, few regions of the pedestrian are visible because of the huge size of occlusion or another person in the image, the pose estimation network $\mc{P}$ would generate heatmaps $\mathbf{M}$ that contain error locations with strong confidence scores. The different orientations of the body also influence the matching results. When optimizing local features generated by our model through the dataset, the above problems often lead to some fatal errors and an unstable training procedure.

Therefore, we intend to aggregate local part features extracted by pose estimation network $\mc{P}$ with the global feature of the whole image to train the model. The main challenge of generating global features is obliging the model to focus on visible parts of the pedestrian as much as possible, while the information of the occlusions occupies a small proportion in features. In the IRM model, for any horizontal region, we have paid attention to visible parts of the pedestrian, but a solitary part hardly represents the whole pedestrian. The relation between visible parts becomes one essential factor for representing the pedestrian. By constructing the relationship between the different parts, we can pass messages to each part which is abstracted as a graph node. The relationship provides a structural understanding between these nodes, and useless features like occlusion features are discarded.

For tackling these problems, we introduce the inter-part relational transformer \cite{attn2017nips}. In the field of natural language processing, the transformer takes sentences or words in some scenes as the input. We could build the internal relationship of various words from the global perspective. But in computer vision, the formation of image data is not constructed in a sequence style. In our proposed inter-part relational transformer, we incorporate transformer encoder architecture into our model. In each partitioning strategy, spatial features $\mathbf{F}_{irm}$ are used to generated different part features. Different partitioned abstract features $\mathbf{X}$ in each group are sent into a weight-sharing transformer unit as shown in \figref{fig:irm} to capture the relationship between them. For the representation of a query part $q$, some value of the feature $\mathbf{X}$ is related to the key part $k$:
\begin{equation}
    \alpha_{q,k}=softmax(\frac{\mathbf{X}{\mathbf{W}_i}^Q(\mathbf{X}{\mathbf{W}_i}^K)^T}{\sqrt{c}}),
\end{equation}
where the attention score $\alpha_{q,k}$ is based on similarities over the query and key part features. $c$ denotes the dimension of the abstract feature $\mathbf{X}$. We can obtain the attention-based features by using a learnable weight $\mathbf{W}^R$ and use multiple projections to generate final features:
\begin{equation}
    \mathbf{\overline{X}}=FFN(\alpha_{q,k}\mathbf{X}\mathbf{W}^{R}),
\end{equation}
where FFN denotes the linear projection on the $\mathbf{X}$ called Feed Forward Network. This module consists of two linear projections with the ReLU activation function and the Dropout function:
\begin{equation}
    FFN(\mathbf{X}) = Dropout(ReLU(\mathbf{X}\mathbf{W}_{1}+\mathbf{b}_{1}))\mathbf{W}_{2} + \mathbf{b}_{2}.
\end{equation}
In this module, partitioned features $\mathbf{X}$ are firstly calculating relation through MHSA module, and output features are then combined with residual features and the layer normalization. The subsequent FFN module would process these features as the same.

\begin{figure}[t]
	\begin{center}
		\includegraphics[width= \linewidth]{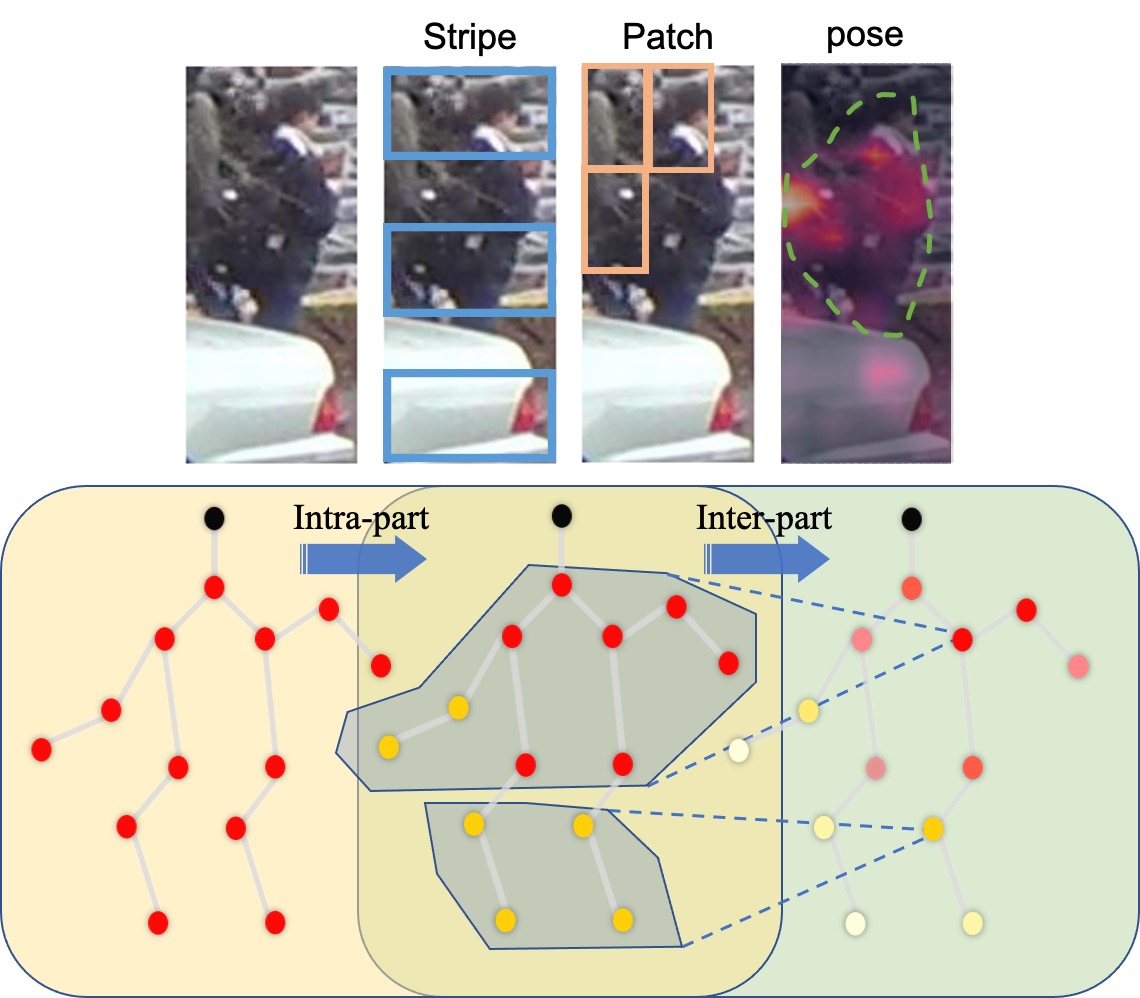}
	\end{center}
	\caption{Different partitioning strategies and the procedure of relation building. Part features learn the self-correlation through intra-part relational module and understand the structural information in each group through inter-part relational transformer.}
	\label{fig:ps}
\end{figure}

\textbf{Partitioning strategy.} Based on the intra-part relation module, there are three strategies for partitioning as shown in \figref{fig:ps}. We firstly apply the stripe pooling which is closely related to the regional pooling to split features $\mathbf{F}_{irm}$ horizontally. We assume features $\mathbf{F}_{irm}$ of each part have already focused on the visible location of the pedestrian, and information of each part still well remains. We mark the striped features as $\mathbf{F}_{stripe} \in \mathbb{R}^{H_f\times 1\times C}$.

The second partitioning strategy divides the spatial features $\mathbf{F}_{irm}$ into patched girds. We assume that the solitary utilization of intra-part relation module and horizontal partitioning strategy loses some information that is incorrectly ignored, resulting in the inability to underlie the thorough relation between their potential information. In the divided part, we aim to find the hidden spatial features and their relation, as a supplement for the global feature. We adopt a GMP layer to select features from $\mathbf{F}_{irm}$ without influencing their potential capability on representation. We mark patched features as $\mathbf{F}_{patch} \in \mathbb{R}^{\frac{H_f}{4}\times \frac{W_f}{2}\times C}$.

For the last partitioning strategy, we incorporate pose heatmaps $\mathbf{M}$ with features $\mathbf{F}_{irm}$ to select the corresponding part feature as shown in \figref{fig:ps}. In addition, we divide all features generated by this strategy into three disjoint groups. We denote the pose-guided features in one of the groups as $\mathbf{F}_{pose} \in \mathbb{R}^{M \times C}$, where $M$ indicates the number of manually divided features.

We apply our inter-part relation transformer to build long-term relationships between partitioned areas. To build relations crossing the part features, we individually send features into our transformer. For each set of local features $\mathbf{F}_{local} \in \{\mathbf{F}_{stripe}, \mathbf{F}_{patch}, \mathbf{F}_{pose} \} $ as mentioned in \secref{sec:1}, we average the first two local features and regard these features as final embedding features $\mathbf{f}_{local}$:
\begin{equation}
    \mathbf{f}_{local} = GAP(IRT(\mathbf{F}_{local})).
\end{equation}
Because the pose estimation network $\mc{P}$ is unstable for confidence score generation due to the domain gap, we propose a confidence score module to generate self-scores $\mathbf{S}_{self}$ and combine with the scores $\mathbf{S}$ mentioned in \secref{sec:1}. The CSM is consistent with a Multi-Layer Perceptron layer, and the $\mathbf{S}_{self}$ is generated as the following:
\begin{equation}
    \mathbf{S}_{self} = ReLU(\mathbf{\hat{F}}_{pose}\mathbf{W}_{1}+\mathbf{b}_{1})\mathbf{W}_{2} + \mathbf{b}_{2},
\end{equation}
where $\mathbf{\hat{F}}_{pose}$ denotes three combined pose-guided features $\mathbf{f}_{pose}$. We use the ReLU activation function to build a non-linear mapping function. $\mathbf{S}_{self}$ is multiplied by $\mathbf{S}$ and a normalization function is applied to smooth the value. The $\mathbf{\widetilde{F}}_{pose}$ is combined with the score at last:
\begin{equation}
    \mathbf{\widetilde{F}}_{pose} = \mathbf{\hat{F}}_{pose} \cdot norm(\mathbf{S}_{self} \cdot \mathbf{S}).
\end{equation}

\textbf{Matching strategy.}
Image retrieval task usually adopts global features and local features to do matching like \cite{delf2017cvprw}. Due to the huge differences between categories, robust local features could well complete traditional retrieval tasks. Different from the image retrieval task, many person re-identification methods use one single global feature with dimension $d$ to represent the person in the image. The main reason is that the discrepancy between local features is not huge, and contributing to insufficient fine-grained details of traditional local features. Therefore the results of matching only based on local features are relatively poor.

On the other hand, occluded person Re-Id often requires local features for precise feature alignment to get better results. Therefore, several methods are proposed to handle this problem. \cite{fpr2019iccv} and \cite{dsr2018cvpr} used spatial feature reconstruction and \cite{horeid2020cvpr} directly used keypoints features. We choose pose-guided local features which contain the fine-grained information of the person. We combine global features and local features to produce results. An initial coarse rank list is established using the global feature, and local features are used to perform more precisely matching on the rank list. With the first ranking list obtained by measuring the global feature, we only conduct local ranking on the top-N images (\eg, N=100) that appear in the list. In nearly all the cases, top-N-selected images cover all the correct results of a query image.

\subsection{Loss}
In the training process, we use cross entropy loss and triplet loss as the supervised classification signals. In every training step, we randomly sample K samples of P pedestrians from the training dataset. For triplet loss, we use hard triplet loss \cite{fpr2019iccv}. The final learning objective of pose-guided local features $\mathbf{\widetilde{F}}_{pose}$ is formulated as the following:
\begin{equation}
    \mathcal{L}_{local} =\frac{1}{p}\sum_{i=1}^p [\mathcal{L}_{cls}(\mathbf{f}^{i}_{pose}) + \mathcal{L}_{tri}(\mathbf{f}^{i}_{pose})],
\end{equation}
where $p$ is the number of pose-guided groups. Learning objective of features $\mathbf{f}_{stripe}$ and $\mathbf{f}_{patch}$ which contain information in the inter- and intra-part manner can be formulated as:
\begin{equation}
    \mc{L}_{global} = \mathcal{L}_{cls}(\mathbf{f}_{stripe}) + \mathcal{L}_{cls}(\mathbf{f}_{patch}),
\end{equation}
The joint loss function is described below:
\begin{equation}
    \mc{L} = \mc{L}_{local} + \mc{L}_{global}.
\end{equation}
With the loss function above, our model can better understand the structural relationship of different part components.

\section{Experiments}

\subsection{Datasets}

\textbf{Occluded-DukeMTMC dataset \cite{pgfa2019iccv}.}  There are 15,618 images for training, 17,661 gallery images and 2,210 occluded query images for testing. Training images contain about 9\% occluded images and the portion of occluded images in the gallery is around 10\%. There always exists one occluded image when computing the distance.

\textbf{Market-1501 dataset \cite{market2015iccv}.}  The dataset includes 32,668 images of 1,501 identities. These pedestrians are captured by six cameras from different scenes and viewpoints. There are 12,936 training images of 751 identities. The testing set contains the remaining images. There is one label to represent the background.

\textbf{DukeMTMC-reID dataset \cite{MTMC2016eccv, duke2017iccv}.} The dataset includes 36,411 images of 1,404 identities. There are eight different cameras to capture the person. This dataset selects 702 identities for training. The remaining images are for testing. In the testing phase, there is only one query image of each person in each camera, and remaining images are reserved for gallery.

\subsection{Implementation details}
\textbf{Model architectures.} For our feature extraction backbone, we use ResNet50-ibn \cite{ibn2018eccv, res2016cvpr} pretrained on ImageNet \cite{imagenet2009cvpr}. We drop its final GAP layer and fully connected layer to extend our module. For pose estimation network, we use HR-Net \cite{hrnet2019cvpr} pretrained on COCO dataset \cite{coco2014eccv} which follows the identical setting as \cite{horeid2020cvpr}. Pose estimation network predicts a total of 17 keypoints, including head, joints of arms, and joints of legs. The parameters of the pose estimation network are set to unchanged during training. We use three stacked transformer units with 512 hidden dimensions in FFN, 0.1 dropout ratio, and the ReLU activation layer. We set a threshold $\tau$ as 0.001 to filter the low confidence score. We use a BN layer to normalize our final embedding features. We use the modified early version of the platform \cite{he2020fastreid} to implement our method.

\textbf{Traning details.} The input images are resized into 384 $\times$ 128. In the training stage, batch size is set to 64 by selecting 16 different identities and 4 samples for each identity. We choose some common data augmentation strategies including padding 10 pixels, random cropping, horizontal flipping, and random erasing \cite{re2020aaai} with a probability of 0.5. During training, Adam optimizer is adopted. We set weight decay $5 \times 10^{-4}$ and we train our model for 60 epochs with initialized learning rate $3.5 \times 10^{-4}$. The learning rate is warmed up for 10 epochs first and decayed by cosine method from the 30th epoch to $1 \times 10^{-6}$. The model is implemented with the pytorch framework and trained with two NVIDIA RTX 2080 Ti.

\textbf{Evaluation metrics.}
For evaluation, cumulative matching characteristic curve and mean average precision (mAP) are adopted.

\begin{table}[]
\caption{Performance (\%) comparisons to the state-of-the-art occluded Re-Id results on the Occluded-DukeMTMC dataset.}
\label{table:occduke}
\begin{tabular}{lcc}
\hline
\multirow{2}{*}{Methods} & \multicolumn{2}{c}{Occluded-DukeMTMC}           \\ \cline{2-3}
                                            & mAP                  & Rank-1               \\ \hline
Part-Aligned \cite{pa2017iccv}               & 20.2                 & 28.8                 \\
PCB \cite{pcb2018eccv}                       & 33.7                 & 42.6                 \\
Adver Occluded \cite{ado2018iccv}            & 32.2                 & 44.5                 \\
FD-GAN \cite{fdgan2018nips}                  & -                    & 40.8                 \\
Part Bilinear \cite{pb2018eccv}              & 36.9                 & -                    \\
PGFA \cite{pgfa2019iccv}                     & 37.3                 & 51.4                 \\
HONet \cite{horeid2020cvpr}                  & 43.8                 & 55.1                 \\
DSR \cite{dsr2018cvpr}                       & 30.4                 & 40.8                 \\ \hline
Baseline (ours)                             & 43.4                 & 51.8                 \\
\textbf{Pirt (ours)}                                 & \textbf{50.9}                 & \textbf{60.0}                \\ \hline
\end{tabular}
\end{table}

\begin{figure*}[t]
	\begin{center}
		\includegraphics[
		width=1 \linewidth,
		]{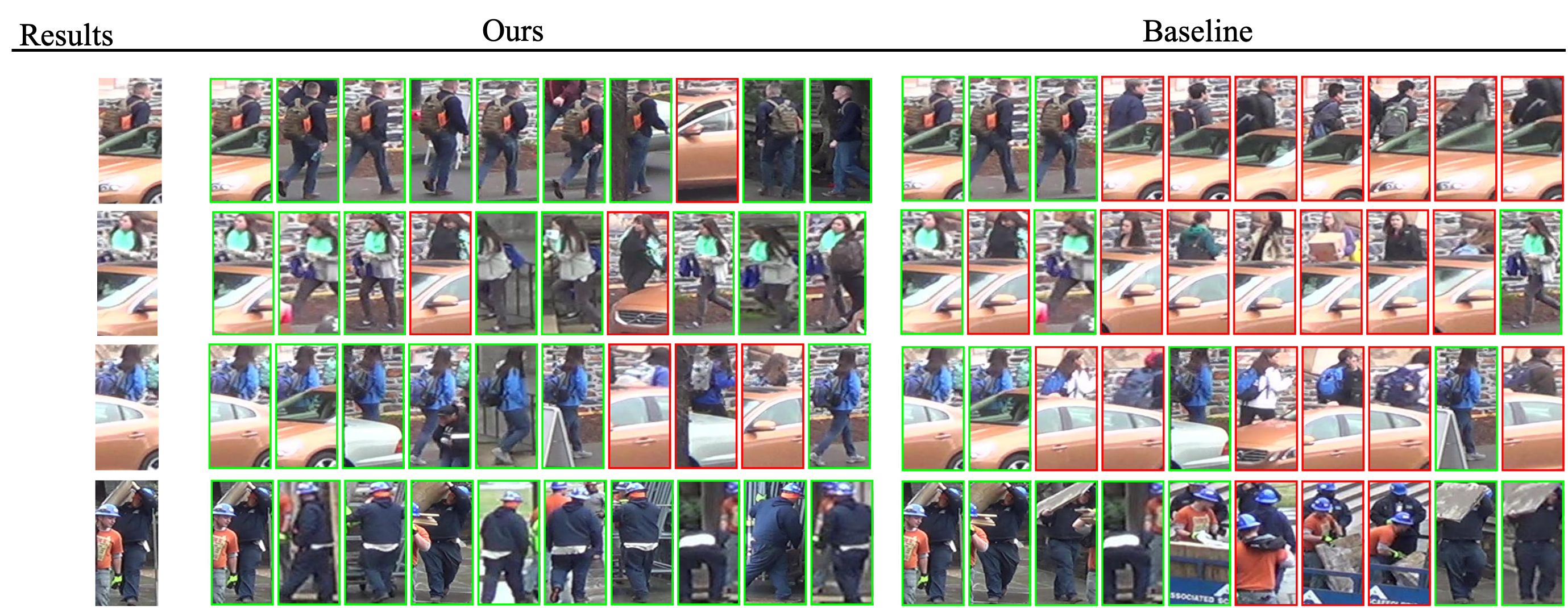}
	\end{center}
	\caption{Comparison of retrieval results. The leftmost denotes query image which is followed by the ten closest matching results (including the same camera). The green color indicates correct retrieval results and the red color indicates error results.}
	\label{fig:vis}
\end{figure*}

\subsection{Experimental Results}

\textbf{Results on the occluded dataset.}
Our experiments are conducted on the occluded Re-Id benchmark dataset with supervised setting, Occluded-DukeMTMC dataset \cite{pgfa2019iccv}. \tabref{table:occduke} shows the comparisons over the dataset Occluded-DukeMTMC. Part-Aligned \cite{pa2017iccv}, PCB \cite{pcb2018eccv} and Adver Occluded \cite{ado2018iccv} are designed for holistic ReID task. FD-GAN \cite{fdgan2018nips}, Part Bilinear \cite{pb2018eccv}, PGFA \cite{pgfa2019iccv}, and HONet \cite{horeid2020cvpr} use the extra pose information. DSR \cite{dsr2018cvpr} does not use extra keypoints detector.

\textbf{Results on holistic datasets.}
If one solution performs well in the occluded dataset, it should also perform well on holistic datasets. To verify whether our method works on holistic datasets, we compare our method with other methods for occluded Re-Id on holistic datasets Market-1501 \cite{market2015iccv} and DukeMTMC-reID \cite{MTMC2016eccv, duke2017iccv}.
\tabref{table:hol} shows that some methods perform well on the Occluded-DukeMTMC dataset, but have different performances on holistic datasets. The method in \cite{fpr2019iccv} can outperform two other methods proposed by \cite{horeid2020cvpr, pgfa2019iccv}, which indicates that simply using external information like keypoints might not achieve the best performance on holistic datasets. We think it is mainly because holistic datasets contain few occlusions, and the ability of these methods on extracting and representing global information from the image to generate pedestrian features is relatively low. Occluded images in the dataset contain massive noise caused by occlusions, and directly using the global feature leads to serious confusion. Semantic information inside the global feature cannot make a good alignment to distinguish corresponding regions in images. In this case, we put forward different partition strategies and attention modules. On the basis of ensuring the robustness of feature representation ability, our method combines local features and global features for more detailed alignment to achieve better results.

\begin{table}[]
\caption{The comparisons over the Market-1501 dataset and the DukeMTMC-reID dataset.}
\label{table:hol}
\begin{tabular}{lcccc}
\hline
\multirow{2}{*}{Methods} & \multicolumn{2}{c}{Market-1501} & \multicolumn{2}{c}{DukeMTMC-reID} \\ \cline{2-5}
                        & mAP            & Rank-1         & mAP             & Rank-1          \\ \hline
PCB \cite{pcb2018eccv}                     & 77.4           & 92.3           & 66.1            & 81.8            \\
DSR \cite{dsr2018cvpr}                     & 64.3          & 83.6          & -               & -               \\
FPR \cite{fpr2019iccv}                     & 86.6          & 95.4          & 78.4           & 88.6           \\
VPM \cite{vpm2019cvpr}                     & 80.8           & 93.0          & 72.6            & 83.6            \\
PGFA \cite{pgfa2019iccv}                    & 76.8           & 91.2           & 65.5            & 82.6            \\
HONet \cite{horeid2020cvpr}                   & 84.9           & 94.2           & 75.6            & 86.9            \\ \hline
\textbf{Pirt (ours)}           & 86.3          & 94.1          & 77.6           & 88.9           \\ \hline
\end{tabular}
\end{table}

In \figref{fig:vis} we make a rank list to compare our baseline with our proposed method. In the first row, our method finds the holistic pedestrian according to visible parts in the query image. In the second and the third row, visible parts are fewer than occlusions thus baseline only capture obvious regions without considering the relation between the person and the occlusion in the image. The last row shows that our method more robustly removes the noise generated by the information of another person. Our method shows the effectiveness to find occluded people.

\begin{table}[t]
\caption{Analysis of three different components: \textbf{P} represent pose generation, \textbf{Intra} represents intra-part relational module and \textbf{Inter} represents inter-part relational transformer.}
\label{table:mc}
\begin{tabular}{c|c|c|c|c}
\hline
\textbf{P}      & \textbf{Intra}     & \textbf{Inter}              & mAP        & Rank-1      \\ \hline
$\times$        & $\times$     & $\times$                  & 43.40      & 51.81       \\
\checkmark      & $\times$     & $\times$                  & 42.23      & 52.17       \\
\checkmark      & \checkmark   & $\times$                  & 47.20      & 55.88       \\
\checkmark      & \checkmark   & \checkmark              & \textbf{50.90}      & \textbf{60.00}       \\ \hline
\end{tabular}
\end{table}

\begin{table}[t]
\caption{The performances on different number of our transformer unit \textbf{N}.}
\label{table:nte}
\begin{tabular}{c|cc}
\hline
\textbf{N} &  mAP     & Rank-1 \\ \hline
1      &  48.93   & 57.51  \\
2      &  49.93   & 59.05  \\
3      &  \textbf{50.90}   & \textbf{60.00}  \\
4      &  50.59   & 59.41  \\ \hline
\end{tabular}
\end{table}

\vspace{20pt}
\subsection{Performance Analysis}

\textbf{Ablations studies.} We evaluate the effectiveness of our proposed method by reconstructing our model to verify the influences of each module. As shown in \tabref{table:mc}, our proposed module improves the performance on the Occluded-DukeMTMC dataset. Baseline method uses only an individual feature vector extracted from the single backbone without using triplet loss function. Then we use HR-Net \cite{hrnet2019cvpr} to extract keypoints from the image and apply averaged local features to present each keypoint. We adopt a GMP and a GAP layer to select representative features as local features in \tabref{table:mc}. For each local feature, we use all the loss functions as described above to constraint the local part feature. But extracting features according to the keypoints from the backbone features hardly leads to a better result. DSR \cite{dsr2018cvpr} shows another type of local features which consists of multiple max poolings with different scales. The final amount of the local features \cite{dsr2018cvpr} is large and the used least square algorithm between two different local features set leads to inefficiency. Our method uses only three robust local features, and the computational complexity is similar to the baseline.

\begin{figure}[t]
	\begin{center}
		\includegraphics[width=0.5 \linewidth]{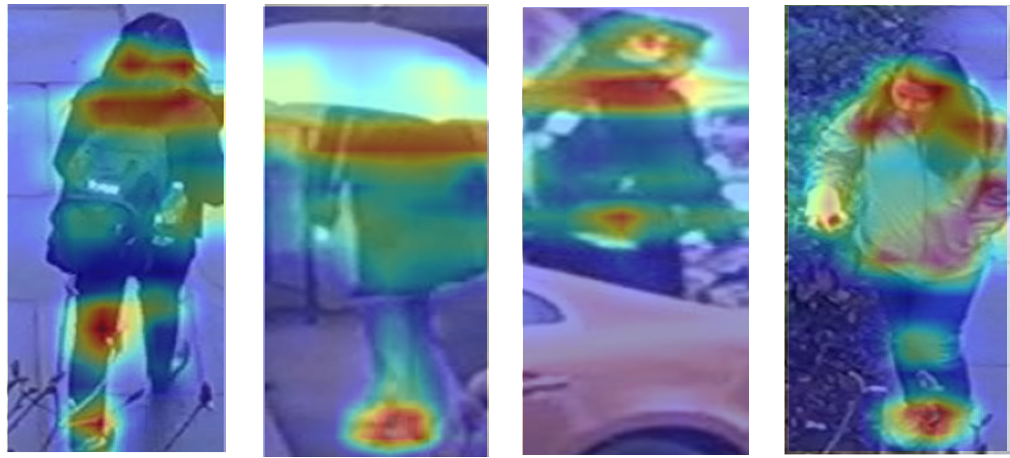}
	\end{center}
	\caption{Visualization on features generated by our proposed intra-part relational module.}
	\label{fig:act}
\end{figure}

\textbf{Effects of inter- and intra-part relational transformer.}  As shown in \tabref{table:mc}, we found that different dividing strategies applied on the spatial features into parts and building the relationship between grids inside each part can give a positive effect to recognize the person in the image as shown in \tabref{table:mc}. Different from \cite{pcb2018eccv}, we do not restrict each striped part using independent cross entropy loss and regard the divided parts as separate local features. With the support of the attention mechanism, merging local features into global features well represents the pedestrian while eliminating lots of parameters. We visualize our model using its weight and activation function as shown in \figref{fig:act}.

\textbf{Effects of the number of our transformer units.} \tabref{table:nte} shows the influence of the different numbers of transformer units on the accuracy. One layer of the transformer unit is insufficient for extracting features. However, the person re-identification task heavily relies on pretrained backbone weights, more encoder layers lead to a tough training procedure while keeping a similar performance. As a result, we set the number of transformer units to 3 to balance the accuracy and the efficiency.

\begin{table}[]
\caption{The effects of different combinations of confidence scores on the matching results. Q and G represent confidence scores of query images and gallery images respectively.}
\label{table:c}
\begin{tabular}{c|cc}
\hline
Combination Methods  & mAP       & Rank-1  \\ \hline
Q \& G              & \textbf{50.90}      & \textbf{60.00}   \\
Q only              & 49.91     & 59.86   \\
G only              & 49.89     & 59.86   \\
No confidence score  & 48.19     & 57.78   \\ \hline
\end{tabular}
\end{table}

\textbf{Effects of confidence score combination method.} Corresponding local features are measured by cosine distance and multiplied with confidence scores generated by confidence score module. There are four situations as shown in \tabref{table:c}. The best performance is conducted in the first situation by simultaneously uniting two confidence scores. The next two situations indicate that using only one of the scores impairs the performances. The reason behind this phenomenon mainly occurred in an occluded scene, \eg, different degrees of occlusion on the corresponding part can get wrong matching results. Ignoring potential confidence score influenced by occlusions may lead to a weak performances.

\section{Conclusions} We propose a novel framework namely Pose-guided inter- and intra-part relational transformers for occluded person Re-Id. In the pose generation stage, our detected keypoints are naturally fused into our module to well represent the holistic object with part groups. We combine these local part features with an attention mechanism to construct long-term correlations and provide a structural understanding of different part components by introducing transformer. With these two complementary relationships constructed from different perspectives, our method reaches the new state of the art on occluded person re-identification task and our experiments also show the effectiveness of our model.

\section{Acknowledgement} This work was supported by grants from National Natural Science Foundation of China (No. 61922006) and CAAI-Huawei MindSpore Open Fund.
\bibliographystyle{ACM-Reference-Format}
\bibliography{sample-base}


\begin{thebibliography}{52}


\ifx \showCODEN    \undefined \def \showCODEN     #1{\unskip}     \fi
\ifx \showDOI      \undefined \def \showDOI       #1{#1}\fi
\ifx \showISBNx    \undefined \def \showISBNx     #1{\unskip}     \fi
\ifx \showISBNxiii \undefined \def \showISBNxiii  #1{\unskip}     \fi
\ifx \showISSN     \undefined \def \showISSN      #1{\unskip}     \fi
\ifx \showLCCN     \undefined \def \showLCCN      #1{\unskip}     \fi
\ifx \shownote     \undefined \def \shownote      #1{#1}          \fi
\ifx \showarticletitle \undefined \def \showarticletitle #1{#1}   \fi
\ifx \showURL      \undefined \def \showURL       {\relax}        \fi
\providecommand\bibfield[2]{#2}
\providecommand\bibinfo[2]{#2}
\providecommand\natexlab[1]{#1}
\providecommand\showeprint[2][]{arXiv:#2}

\bibitem[\protect\citeauthoryear{Cao, Araujo, and Sim}{Cao
  et~al\mbox{.}}{2020}]%
        {delg2020eccv}
\bibfield{author}{\bibinfo{person}{Bingyi Cao}, \bibinfo{person}{Andr{\'e}
  Araujo}, {and} \bibinfo{person}{Jack Sim}.} \bibinfo{year}{2020}\natexlab{}.
\newblock \showarticletitle{Unifying deep local and global features for image
  search}. In \bibinfo{booktitle}{\emph{European Conference on Computer
  Vision}}. Springer, \bibinfo{pages}{726--743}.
\newblock


\bibitem[\protect\citeauthoryear{Chen, Deng, and Hu}{Chen
  et~al\mbox{.}}{2019}]%
        {mixedo2019iccv}
\bibfield{author}{\bibinfo{person}{Binghui Chen}, \bibinfo{person}{Weihong
  Deng}, {and} \bibinfo{person}{Jiani Hu}.} \bibinfo{year}{2019}\natexlab{}.
\newblock \showarticletitle{Mixed high-order attention network for person
  re-identification}. In \bibinfo{booktitle}{\emph{Proceedings of the IEEE/CVF
  International Conference on Computer Vision}}. \bibinfo{pages}{371--381}.
\newblock


\bibitem[\protect\citeauthoryear{Chen, Fu, Zhao, Zheng, Song, Ji, and
  Yang}{Chen et~al\mbox{.}}{2020}]%
        {sgsnet2020cvpr}
\bibfield{author}{\bibinfo{person}{Xuesong Chen}, \bibinfo{person}{Canmiao Fu},
  \bibinfo{person}{Yong Zhao}, \bibinfo{person}{Feng Zheng},
  \bibinfo{person}{Jingkuan Song}, \bibinfo{person}{Rongrong Ji}, {and}
  \bibinfo{person}{Yi Yang}.} \bibinfo{year}{2020}\natexlab{}.
\newblock \showarticletitle{Salience-guided cascaded suppression network for
  person re-identification}. In \bibinfo{booktitle}{\emph{Proceedings of the
  IEEE/CVF Conference on Computer Vision and Pattern Recognition}}.
  \bibinfo{pages}{3300--3310}.
\newblock


\bibitem[\protect\citeauthoryear{Dai, Chen, Gu, Zhu, and Tan}{Dai
  et~al\mbox{.}}{2019}]%
        {bdbnet2019iccv}
\bibfield{author}{\bibinfo{person}{Zuozhuo Dai}, \bibinfo{person}{Mingqiang
  Chen}, \bibinfo{person}{Xiaodong Gu}, \bibinfo{person}{Siyu Zhu}, {and}
  \bibinfo{person}{Ping Tan}.} \bibinfo{year}{2019}\natexlab{}.
\newblock \showarticletitle{Batch dropblock network for person
  re-identification and beyond}. In \bibinfo{booktitle}{\emph{Proceedings of
  the IEEE/CVF International Conference on Computer Vision}}.
  \bibinfo{pages}{3691--3701}.
\newblock


\bibitem[\protect\citeauthoryear{Deng, Dong, Socher, Li, Li, and Fei-Fei}{Deng
  et~al\mbox{.}}{2009}]%
        {imagenet2009cvpr}
\bibfield{author}{\bibinfo{person}{Jia Deng}, \bibinfo{person}{Wei Dong},
  \bibinfo{person}{Richard Socher}, \bibinfo{person}{Li-Jia Li},
  \bibinfo{person}{Kai Li}, {and} \bibinfo{person}{Li Fei-Fei}.}
  \bibinfo{year}{2009}\natexlab{}.
\newblock \showarticletitle{Imagenet: A large-scale hierarchical image
  database}. In \bibinfo{booktitle}{\emph{IEEE Conference on Computer Vision
  and Pattern Recognition (CVPR)}}. \bibinfo{pages}{248--255}.
\newblock


\bibitem[\protect\citeauthoryear{DeTone, Malisiewicz, and Rabinovich}{DeTone
  et~al\mbox{.}}{2018}]%
        {superp2018iccvw}
\bibfield{author}{\bibinfo{person}{Daniel DeTone}, \bibinfo{person}{Tomasz
  Malisiewicz}, {and} \bibinfo{person}{Andrew Rabinovich}.}
  \bibinfo{year}{2018}\natexlab{}.
\newblock \showarticletitle{Superpoint: Self-supervised interest point
  detection and description}. In \bibinfo{booktitle}{\emph{Proceedings of the
  IEEE conference on computer vision and pattern recognition workshops}}.
  \bibinfo{pages}{224--236}.
\newblock


\bibitem[\protect\citeauthoryear{Devlin, Chang, Lee, and Toutanova}{Devlin
  et~al\mbox{.}}{2019}]%
        {bert2018naacl}
\bibfield{author}{\bibinfo{person}{Jacob Devlin}, \bibinfo{person}{Ming-Wei
  Chang}, \bibinfo{person}{Kenton Lee}, {and} \bibinfo{person}{Kristina
  Toutanova}.} \bibinfo{year}{2019}\natexlab{}.
\newblock \showarticletitle{BERT: Pre-training of Deep Bidirectional
  Transformers for Language Understanding.}. In
  \bibinfo{booktitle}{\emph{NAACL-HLT (1)}},
  \bibfield{editor}{\bibinfo{person}{Jill Burstein}, \bibinfo{person}{Christy
  Doran}, {and} \bibinfo{person}{Thamar Solorio}} (Eds.).
  \bibinfo{publisher}{Association for Computational Linguistics},
  \bibinfo{pages}{4171--4186}.
\newblock
\showISBNx{978-1-950737-13-0}
\urldef\tempurl%
\url{http://dblp.uni-trier.de/db/conf/naacl/naacl2019-1.html#DevlinCLT19}
\showURL{%
\tempurl}


\bibitem[\protect\citeauthoryear{Gao, Wang, Lu, and Liu}{Gao
  et~al\mbox{.}}{2020}]%
        {pvpm2020cvpr}
\bibfield{author}{\bibinfo{person}{Shang Gao}, \bibinfo{person}{Jingya Wang},
  \bibinfo{person}{Huchuan Lu}, {and} \bibinfo{person}{Zimo Liu}.}
  \bibinfo{year}{2020}\natexlab{}.
\newblock \showarticletitle{Pose-guided visible part matching for occluded
  person ReID}. In \bibinfo{booktitle}{\emph{Proceedings of the IEEE/CVF
  Conference on Computer Vision and Pattern Recognition}}.
  \bibinfo{pages}{11744--11752}.
\newblock


\bibitem[\protect\citeauthoryear{Ge, Li, Zhao, Yin, Yi, Wang, and Li}{Ge
  et~al\mbox{.}}{2018}]%
        {fdgan2018nips}
\bibfield{author}{\bibinfo{person}{Yixiao Ge}, \bibinfo{person}{Zhuowan Li},
  \bibinfo{person}{Haiyu Zhao}, \bibinfo{person}{Guojun Yin},
  \bibinfo{person}{Shuai Yi}, \bibinfo{person}{Xiaogang Wang}, {and}
  \bibinfo{person}{Hongsheng Li}.} \bibinfo{year}{2018}\natexlab{}.
\newblock \showarticletitle{FD-GAN: Pose-guided Feature Distilling GAN for
  Robust Person Re-identification}. In \bibinfo{booktitle}{\emph{Advances in
  Neural Information Processing Systems}}. \bibinfo{pages}{1229--1240}.
\newblock


\bibitem[\protect\citeauthoryear{Guo, Yuan, Huang, Zhang, Yao, and Han}{Guo
  et~al\mbox{.}}{2019}]%
        {dual2019iccv}
\bibfield{author}{\bibinfo{person}{Jianyuan Guo}, \bibinfo{person}{Yuhui Yuan},
  \bibinfo{person}{Lang Huang}, \bibinfo{person}{Chao Zhang},
  \bibinfo{person}{Jin-Ge Yao}, {and} \bibinfo{person}{Kai Han}.}
  \bibinfo{year}{2019}\natexlab{}.
\newblock \showarticletitle{Beyond human parts: Dual part-aligned
  representations for person re-identification}. In
  \bibinfo{booktitle}{\emph{Proceedings of the IEEE/CVF International
  Conference on Computer Vision}}. \bibinfo{pages}{3642--3651}.
\newblock


\bibitem[\protect\citeauthoryear{He, Li, Zhao, and Tian}{He
  et~al\mbox{.}}{2019a}]%
        {he2019part}
\bibfield{author}{\bibinfo{person}{Bing He}, \bibinfo{person}{Jia Li},
  \bibinfo{person}{Yifan Zhao}, {and} \bibinfo{person}{Yonghong Tian}.}
  \bibinfo{year}{2019}\natexlab{a}.
\newblock \showarticletitle{Part-regularized near-duplicate vehicle
  re-identification}. In \bibinfo{booktitle}{\emph{Proceedings of the IEEE/CVF
  Conference on Computer Vision and Pattern Recognition}}.
  \bibinfo{pages}{3997--4005}.
\newblock


\bibitem[\protect\citeauthoryear{He, Zhang, Ren, and Sun}{He
  et~al\mbox{.}}{2016}]%
        {res2016cvpr}
\bibfield{author}{\bibinfo{person}{Kaiming He}, \bibinfo{person}{Xiangyu
  Zhang}, \bibinfo{person}{Shaoqing Ren}, {and} \bibinfo{person}{Jian Sun}.}
  \bibinfo{year}{2016}\natexlab{}.
\newblock \showarticletitle{Deep residual learning for image recognition}. In
  \bibinfo{booktitle}{\emph{IEEE Conference on Computer Vision and Pattern
  Recognition (CVPR)}}. \bibinfo{pages}{770--778}.
\newblock


\bibitem[\protect\citeauthoryear{He, Liang, Li, and Sun}{He
  et~al\mbox{.}}{2018}]%
        {dsr2018cvpr}
\bibfield{author}{\bibinfo{person}{Lingxiao He}, \bibinfo{person}{Jian Liang},
  \bibinfo{person}{Haiqing Li}, {and} \bibinfo{person}{Zhenan Sun}.}
  \bibinfo{year}{2018}\natexlab{}.
\newblock \showarticletitle{Deep spatial feature reconstruction for partial
  person re-identification: Alignment-free approach}. In
  \bibinfo{booktitle}{\emph{Proceedings of the IEEE Conference on Computer
  Vision and Pattern Recognition}}. \bibinfo{pages}{7073--7082}.
\newblock


\bibitem[\protect\citeauthoryear{He, Liao, Liu, Liu, Cheng, and Mei}{He
  et~al\mbox{.}}{2020}]%
        {he2020fastreid}
\bibfield{author}{\bibinfo{person}{Lingxiao He}, \bibinfo{person}{Xingyu Liao},
  \bibinfo{person}{Wu Liu}, \bibinfo{person}{Xinchen Liu},
  \bibinfo{person}{Peng Cheng}, {and} \bibinfo{person}{Tao Mei}.}
  \bibinfo{year}{2020}\natexlab{}.
\newblock \showarticletitle{FastReID: A Pytorch Toolbox for General Instance
  Re-identification}.
\newblock \bibinfo{journal}{\emph{arXiv preprint arXiv:2006.02631}}
  (\bibinfo{year}{2020}).
\newblock


\bibitem[\protect\citeauthoryear{He and Liu}{He and Liu}{2020}]%
        {gsfl2020eccv}
\bibfield{author}{\bibinfo{person}{Lingxiao He} {and} \bibinfo{person}{Wu
  Liu}.} \bibinfo{year}{2020}\natexlab{}.
\newblock \showarticletitle{Guided Saliency Feature Learning for Person
  Re-identification in Crowded Scenes}. In \bibinfo{booktitle}{\emph{European
  Conference on Computer Vision}}. Springer, \bibinfo{pages}{357--373}.
\newblock


\bibitem[\protect\citeauthoryear{He, Wang, Liu, Zhao, Sun, and Feng}{He
  et~al\mbox{.}}{2019b}]%
        {fpr2019iccv}
\bibfield{author}{\bibinfo{person}{Lingxiao He}, \bibinfo{person}{Yinggang
  Wang}, \bibinfo{person}{Wu Liu}, \bibinfo{person}{He Zhao},
  \bibinfo{person}{Zhenan Sun}, {and} \bibinfo{person}{Jiashi Feng}.}
  \bibinfo{year}{2019}\natexlab{b}.
\newblock \showarticletitle{Foreground-aware pyramid reconstruction for
  alignment-free occluded person re-identification}. In
  \bibinfo{booktitle}{\emph{Proceedings of the IEEE/CVF International
  Conference on Computer Vision}}. \bibinfo{pages}{8450--8459}.
\newblock


\bibitem[\protect\citeauthoryear{Hou, Ma, Chang, Gu, Shan, and Chen}{Hou
  et~al\mbox{.}}{2019}]%
        {ianet2019cvpr}
\bibfield{author}{\bibinfo{person}{Ruibing Hou}, \bibinfo{person}{Bingpeng Ma},
  \bibinfo{person}{Hong Chang}, \bibinfo{person}{Xinqian Gu},
  \bibinfo{person}{Shiguang Shan}, {and} \bibinfo{person}{Xilin Chen}.}
  \bibinfo{year}{2019}\natexlab{}.
\newblock \showarticletitle{Interaction-and-aggregation network for person
  re-identification}. In \bibinfo{booktitle}{\emph{Proceedings of the IEEE/CVF
  Conference on Computer Vision and Pattern Recognition}}.
  \bibinfo{pages}{9317--9326}.
\newblock


\bibitem[\protect\citeauthoryear{Huang, Li, Zhang, Chen, and Huang}{Huang
  et~al\mbox{.}}{2018}]%
        {ado2018iccv}
\bibfield{author}{\bibinfo{person}{Houjing Huang}, \bibinfo{person}{Dangwei
  Li}, \bibinfo{person}{Zhang Zhang}, \bibinfo{person}{Xiaotang Chen}, {and}
  \bibinfo{person}{Kaiqi Huang}.} \bibinfo{year}{2018}\natexlab{}.
\newblock \showarticletitle{Adversarially occluded samples for person
  re-identification}. In \bibinfo{booktitle}{\emph{Proceedings of the IEEE
  Conference on Computer Vision and Pattern Recognition}}.
  \bibinfo{pages}{5098--5107}.
\newblock


\bibitem[\protect\citeauthoryear{Lan, Zhu, Hauptmann, and Newsam}{Lan
  et~al\mbox{.}}{2017}]%
        {delf2017cvprw}
\bibfield{author}{\bibinfo{person}{Zhenzhong Lan}, \bibinfo{person}{Yi Zhu},
  \bibinfo{person}{Alexander~G Hauptmann}, {and} \bibinfo{person}{Shawn
  Newsam}.} \bibinfo{year}{2017}\natexlab{}.
\newblock \showarticletitle{Deep local video feature for action recognition}.
  In \bibinfo{booktitle}{\emph{Proceedings of the IEEE conference on computer
  vision and pattern recognition workshops}}. \bibinfo{pages}{1--7}.
\newblock


\bibitem[\protect\citeauthoryear{Lin, Maire, Belongie, Hays, Perona, Ramanan,
  Doll{\'a}r, and Zitnick}{Lin et~al\mbox{.}}{2014}]%
        {coco2014eccv}
\bibfield{author}{\bibinfo{person}{Tsung-Yi Lin}, \bibinfo{person}{Michael
  Maire}, \bibinfo{person}{Serge Belongie}, \bibinfo{person}{James Hays},
  \bibinfo{person}{Pietro Perona}, \bibinfo{person}{Deva Ramanan},
  \bibinfo{person}{Piotr Doll{\'a}r}, {and} \bibinfo{person}{C~Lawrence
  Zitnick}.} \bibinfo{year}{2014}\natexlab{}.
\newblock \showarticletitle{Microsoft coco: Common objects in context}. In
  \bibinfo{booktitle}{\emph{European conference on computer vision}}. Springer,
  \bibinfo{pages}{740--755}.
\newblock


\bibitem[\protect\citeauthoryear{Miao, Wu, Liu, Ding, and Yang}{Miao
  et~al\mbox{.}}{2019}]%
        {pgfa2019iccv}
\bibfield{author}{\bibinfo{person}{Jiaxu Miao}, \bibinfo{person}{Yu Wu},
  \bibinfo{person}{Ping Liu}, \bibinfo{person}{Yuhang Ding}, {and}
  \bibinfo{person}{Yi Yang}.} \bibinfo{year}{2019}\natexlab{}.
\newblock \showarticletitle{Pose-guided feature alignment for occluded person
  re-identification}. In \bibinfo{booktitle}{\emph{Proceedings of the IEEE/CVF
  International Conference on Computer Vision}}. \bibinfo{pages}{542--551}.
\newblock


\bibitem[\protect\citeauthoryear{Pan, Luo, Shi, and Tang}{Pan
  et~al\mbox{.}}{2018}]%
        {ibn2018eccv}
\bibfield{author}{\bibinfo{person}{Xingang Pan}, \bibinfo{person}{Ping Luo},
  \bibinfo{person}{Jianping Shi}, {and} \bibinfo{person}{Xiaoou Tang}.}
  \bibinfo{year}{2018}\natexlab{}.
\newblock \showarticletitle{Two at Once: Enhancing Learning and Generalization
  Capacities via IBN-Net}. In \bibinfo{booktitle}{\emph{Proceedings of the
  European Conference on Computer Vision (ECCV)}}.
\newblock


\bibitem[\protect\citeauthoryear{Parmar, Vaswani, Uszkoreit, Kaiser, Shazeer,
  Ku, and Tran}{Parmar et~al\mbox{.}}{2018}]%
        {imgtrans2018icml}
\bibfield{author}{\bibinfo{person}{Niki Parmar}, \bibinfo{person}{Ashish
  Vaswani}, \bibinfo{person}{Jakob Uszkoreit}, \bibinfo{person}{Lukasz Kaiser},
  \bibinfo{person}{Noam Shazeer}, \bibinfo{person}{Alexander Ku}, {and}
  \bibinfo{person}{Dustin Tran}.} \bibinfo{year}{2018}\natexlab{}.
\newblock \showarticletitle{Image transformer}. In
  \bibinfo{booktitle}{\emph{International Conference on Machine Learning}}.
  PMLR, \bibinfo{pages}{4055--4064}.
\newblock


\bibitem[\protect\citeauthoryear{Quan, Dong, Wu, Zhu, and Yang}{Quan
  et~al\mbox{.}}{2019}]%
        {auto2019iccv}
\bibfield{author}{\bibinfo{person}{Ruijie Quan}, \bibinfo{person}{Xuanyi Dong},
  \bibinfo{person}{Yu Wu}, \bibinfo{person}{Linchao Zhu}, {and}
  \bibinfo{person}{Yi Yang}.} \bibinfo{year}{2019}\natexlab{}.
\newblock \showarticletitle{Auto-reid: Searching for a part-aware convnet for
  person re-identification}. In \bibinfo{booktitle}{\emph{Proceedings of the
  IEEE/CVF International Conference on Computer Vision}}.
  \bibinfo{pages}{3750--3759}.
\newblock


\bibitem[\protect\citeauthoryear{Ristani, Solera, Zou, Cucchiara, and
  Tomasi}{Ristani et~al\mbox{.}}{2016}]%
        {MTMC2016eccv}
\bibfield{author}{\bibinfo{person}{Ergys Ristani}, \bibinfo{person}{Francesco
  Solera}, \bibinfo{person}{Roger Zou}, \bibinfo{person}{Rita Cucchiara}, {and}
  \bibinfo{person}{Carlo Tomasi}.} \bibinfo{year}{2016}\natexlab{}.
\newblock \showarticletitle{Performance measures and a data set for
  multi-target, multi-camera tracking}. In \bibinfo{booktitle}{\emph{European
  conference on computer vision}}. Springer, \bibinfo{pages}{17--35}.
\newblock


\bibitem[\protect\citeauthoryear{Sarlin, DeTone, Malisiewicz, and
  Rabinovich}{Sarlin et~al\mbox{.}}{2020}]%
        {superg2020cvpr}
\bibfield{author}{\bibinfo{person}{Paul-Edouard Sarlin},
  \bibinfo{person}{Daniel DeTone}, \bibinfo{person}{Tomasz Malisiewicz}, {and}
  \bibinfo{person}{Andrew Rabinovich}.} \bibinfo{year}{2020}\natexlab{}.
\newblock \showarticletitle{Superglue: Learning feature matching with graph
  neural networks}. In \bibinfo{booktitle}{\emph{Proceedings of the IEEE/CVF
  conference on computer vision and pattern recognition}}.
  \bibinfo{pages}{4938--4947}.
\newblock


\bibitem[\protect\citeauthoryear{Song, Huang, Ouyang, and Wang}{Song
  et~al\mbox{.}}{2018}]%
        {mgcam2018cvpr}
\bibfield{author}{\bibinfo{person}{Chunfeng Song}, \bibinfo{person}{Yan Huang},
  \bibinfo{person}{Wanli Ouyang}, {and} \bibinfo{person}{Liang Wang}.}
  \bibinfo{year}{2018}\natexlab{}.
\newblock \showarticletitle{Mask-guided contrastive attention model for person
  re-identification}. In \bibinfo{booktitle}{\emph{Proceedings of the IEEE
  Conference on Computer Vision and Pattern Recognition}}.
  \bibinfo{pages}{1179--1188}.
\newblock


\bibitem[\protect\citeauthoryear{Su, Li, Zhang, Xing, Gao, and Tian}{Su
  et~al\mbox{.}}{2017}]%
        {posed2017iccv}
\bibfield{author}{\bibinfo{person}{Chi Su}, \bibinfo{person}{Jianing Li},
  \bibinfo{person}{Shiliang Zhang}, \bibinfo{person}{Junliang Xing},
  \bibinfo{person}{Wen Gao}, {and} \bibinfo{person}{Qi Tian}.}
  \bibinfo{year}{2017}\natexlab{}.
\newblock \showarticletitle{Pose-driven deep convolutional model for person
  re-identification}. In \bibinfo{booktitle}{\emph{Proceedings of the IEEE
  international conference on computer vision}}. \bibinfo{pages}{3960--3969}.
\newblock


\bibitem[\protect\citeauthoryear{Suh, Wang, Tang, Mei, and Lee}{Suh
  et~al\mbox{.}}{2018}]%
        {pb2018eccv}
\bibfield{author}{\bibinfo{person}{Yumin Suh}, \bibinfo{person}{Jingdong Wang},
  \bibinfo{person}{Siyu Tang}, \bibinfo{person}{Tao Mei}, {and}
  \bibinfo{person}{Kyoung~Mu Lee}.} \bibinfo{year}{2018}\natexlab{}.
\newblock \showarticletitle{Part-aligned bilinear representations for person
  re-identification}. In \bibinfo{booktitle}{\emph{Proceedings of the European
  Conference on Computer Vision (ECCV)}}. \bibinfo{pages}{402--419}.
\newblock


\bibitem[\protect\citeauthoryear{Sun, Xiao, Liu, and Wang}{Sun
  et~al\mbox{.}}{2019a}]%
        {hrnet2019cvpr}
\bibfield{author}{\bibinfo{person}{Ke Sun}, \bibinfo{person}{Bin Xiao},
  \bibinfo{person}{Dong Liu}, {and} \bibinfo{person}{Jingdong Wang}.}
  \bibinfo{year}{2019}\natexlab{a}.
\newblock \showarticletitle{Deep high-resolution representation learning for
  human pose estimation}. In \bibinfo{booktitle}{\emph{Proceedings of the
  IEEE/CVF Conference on Computer Vision and Pattern Recognition}}.
  \bibinfo{pages}{5693--5703}.
\newblock


\bibitem[\protect\citeauthoryear{Sun, Xu, Li, Zhang, Li, Wang, and Sun}{Sun
  et~al\mbox{.}}{2019b}]%
        {vpm2019cvpr}
\bibfield{author}{\bibinfo{person}{Yifan Sun}, \bibinfo{person}{Qin Xu},
  \bibinfo{person}{Yali Li}, \bibinfo{person}{Chi Zhang},
  \bibinfo{person}{Yikang Li}, \bibinfo{person}{Shengjin Wang}, {and}
  \bibinfo{person}{Jian Sun}.} \bibinfo{year}{2019}\natexlab{b}.
\newblock \showarticletitle{Perceive where to focus: Learning visibility-aware
  part-level features for partial person re-identification}. In
  \bibinfo{booktitle}{\emph{Proceedings of the IEEE/CVF Conference on Computer
  Vision and Pattern Recognition}}. \bibinfo{pages}{393--402}.
\newblock


\bibitem[\protect\citeauthoryear{Sun, Zheng, Yang, Tian, and Wang}{Sun
  et~al\mbox{.}}{2018}]%
        {pcb2018eccv}
\bibfield{author}{\bibinfo{person}{Yifan Sun}, \bibinfo{person}{Liang Zheng},
  \bibinfo{person}{Yi Yang}, \bibinfo{person}{Qi Tian}, {and}
  \bibinfo{person}{Shengjin Wang}.} \bibinfo{year}{2018}\natexlab{}.
\newblock \showarticletitle{Beyond part models: Person retrieval with refined
  part pooling (and a strong convolutional baseline)}. In
  \bibinfo{booktitle}{\emph{Proceedings of the European conference on computer
  vision (ECCV)}}. \bibinfo{pages}{480--496}.
\newblock


\bibitem[\protect\citeauthoryear{Tay, Roy, and Yap}{Tay et~al\mbox{.}}{2019}]%
        {aanet2019cvpr}
\bibfield{author}{\bibinfo{person}{Chiat-Pin Tay}, \bibinfo{person}{Sharmili
  Roy}, {and} \bibinfo{person}{Kim-Hui Yap}.} \bibinfo{year}{2019}\natexlab{}.
\newblock \showarticletitle{Aanet: Attribute attention network for person
  re-identifications}. In \bibinfo{booktitle}{\emph{Proceedings of the IEEE/CVF
  Conference on Computer Vision and Pattern Recognition}}.
  \bibinfo{pages}{7134--7143}.
\newblock


\bibitem[\protect\citeauthoryear{Vaswani, Shazeer, Parmar, Uszkoreit, Jones,
  Gomez, Kaiser, and Polosukhin}{Vaswani et~al\mbox{.}}{2017}]%
        {attn2017nips}
\bibfield{author}{\bibinfo{person}{Ashish Vaswani}, \bibinfo{person}{Noam
  Shazeer}, \bibinfo{person}{Niki Parmar}, \bibinfo{person}{Jakob Uszkoreit},
  \bibinfo{person}{Llion Jones}, \bibinfo{person}{Aidan~N Gomez},
  \bibinfo{person}{\L~ukasz Kaiser}, {and} \bibinfo{person}{Illia Polosukhin}.}
  \bibinfo{year}{2017}\natexlab{}.
\newblock \showarticletitle{Attention is All you Need}. In
  \bibinfo{booktitle}{\emph{Advances in Neural Information Processing
  Systems}}, \bibfield{editor}{\bibinfo{person}{I.~Guyon},
  \bibinfo{person}{U.~V. Luxburg}, \bibinfo{person}{S.~Bengio},
  \bibinfo{person}{H.~Wallach}, \bibinfo{person}{R.~Fergus},
  \bibinfo{person}{S.~Vishwanathan}, {and} \bibinfo{person}{R.~Garnett}}
  (Eds.), Vol.~\bibinfo{volume}{30}. \bibinfo{publisher}{Curran Associates,
  Inc.}
\newblock
\urldef\tempurl%
\url{https://proceedings.neurips.cc/paper/2017/file/3f5ee243547dee91fbd053c1c4a845aa-Paper.pdf}
\showURL{%
\tempurl}


\bibitem[\protect\citeauthoryear{Wang, Yang, Liu, Wang, Yang, Wang, Yu, Zhou,
  and Sun}{Wang et~al\mbox{.}}{2020}]%
        {horeid2020cvpr}
\bibfield{author}{\bibinfo{person}{Guan'an Wang}, \bibinfo{person}{Shuo Yang},
  \bibinfo{person}{Huanyu Liu}, \bibinfo{person}{Zhicheng Wang},
  \bibinfo{person}{Yang Yang}, \bibinfo{person}{Shuliang Wang},
  \bibinfo{person}{Gang Yu}, \bibinfo{person}{Erjin Zhou}, {and}
  \bibinfo{person}{Jian Sun}.} \bibinfo{year}{2020}\natexlab{}.
\newblock \showarticletitle{High-order information matters: Learning relation
  and topology for occluded person re-identification}. In
  \bibinfo{booktitle}{\emph{Proceedings of the IEEE/CVF Conference on Computer
  Vision and Pattern Recognition}}. \bibinfo{pages}{6449--6458}.
\newblock


\bibitem[\protect\citeauthoryear{Wang, Yan, and Yang}{Wang
  et~al\mbox{.}}{2019}]%
        {cgm2019iccv}
\bibfield{author}{\bibinfo{person}{Runzhong Wang}, \bibinfo{person}{Junchi
  Yan}, {and} \bibinfo{person}{Xiaokang Yang}.}
  \bibinfo{year}{2019}\natexlab{}.
\newblock \showarticletitle{Learning combinatorial embedding networks for deep
  graph matching}. In \bibinfo{booktitle}{\emph{Proceedings of the IEEE/CVF
  International Conference on Computer Vision}}. \bibinfo{pages}{3056--3065}.
\newblock


\bibitem[\protect\citeauthoryear{Wei, Zhang, Gao, and Tian}{Wei
  et~al\mbox{.}}{2018}]%
        {msmt172018cvpr}
\bibfield{author}{\bibinfo{person}{Longhui Wei}, \bibinfo{person}{Shiliang
  Zhang}, \bibinfo{person}{Wen Gao}, {and} \bibinfo{person}{Qi Tian}.}
  \bibinfo{year}{2018}\natexlab{}.
\newblock \showarticletitle{Person transfer gan to bridge domain gap for person
  re-identification}. In \bibinfo{booktitle}{\emph{Proceedings of the IEEE
  conference on computer vision and pattern recognition}}.
  \bibinfo{pages}{79--88}.
\newblock


\bibitem[\protect\citeauthoryear{Xia, Gong, Zhang, and Poellabauer}{Xia
  et~al\mbox{.}}{2019}]%
        {son2019iccv}
\bibfield{author}{\bibinfo{person}{Bryan~Ning Xia}, \bibinfo{person}{Yuan
  Gong}, \bibinfo{person}{Yizhe Zhang}, {and} \bibinfo{person}{Christian
  Poellabauer}.} \bibinfo{year}{2019}\natexlab{}.
\newblock \showarticletitle{Second-order non-local attention networks for
  person re-identification}. In \bibinfo{booktitle}{\emph{Proceedings of the
  IEEE/CVF International Conference on Computer Vision}}.
  \bibinfo{pages}{3760--3769}.
\newblock


\bibitem[\protect\citeauthoryear{{Ye}, {Shen}, {Lin}, {Xiang}, {Shao}, and
  {Hoi}}{{Ye} et~al\mbox{.}}{2021}]%
        {reidsurvey-tpami}
\bibfield{author}{\bibinfo{person}{M. {Ye}}, \bibinfo{person}{J. {Shen}},
  \bibinfo{person}{G. {Lin}}, \bibinfo{person}{T. {Xiang}}, \bibinfo{person}{L.
  {Shao}}, {and} \bibinfo{person}{S.~C.~H. {Hoi}}.}
  \bibinfo{year}{2021}\natexlab{}.
\newblock \showarticletitle{Deep Learning for Person Re-identification: A
  Survey and Outlook}.
\newblock \bibinfo{journal}{\emph{IEEE Transactions on Pattern Analysis and
  Machine Intelligence}} (\bibinfo{year}{2021}), \bibinfo{pages}{1--1}.
\newblock
\urldef\tempurl%
\url{https://doi.org/10.1109/TPAMI.2021.3054775}
\showDOI{\tempurl}


\bibitem[\protect\citeauthoryear{Zanfir and Sminchisescu}{Zanfir and
  Sminchisescu}{2018}]%
        {dgm2018cvpr}
\bibfield{author}{\bibinfo{person}{Andrei Zanfir} {and}
  \bibinfo{person}{Cristian Sminchisescu}.} \bibinfo{year}{2018}\natexlab{}.
\newblock \showarticletitle{Deep learning of graph matching}. In
  \bibinfo{booktitle}{\emph{Proceedings of the IEEE conference on computer
  vision and pattern recognition}}. \bibinfo{pages}{2684--2693}.
\newblock


\bibitem[\protect\citeauthoryear{Zhang, Lan, Zeng, Jin, and Chen}{Zhang
  et~al\mbox{.}}{2020}]%
        {raga2020cvpr}
\bibfield{author}{\bibinfo{person}{Zhizheng Zhang}, \bibinfo{person}{Cuiling
  Lan}, \bibinfo{person}{Wenjun Zeng}, \bibinfo{person}{Xin Jin}, {and}
  \bibinfo{person}{Zhibo Chen}.} \bibinfo{year}{2020}\natexlab{}.
\newblock \showarticletitle{Relation-aware global attention for person
  re-identification}. In \bibinfo{booktitle}{\emph{Proceedings of the IEEE/CVF
  Conference on Computer Vision and Pattern Recognition}}.
  \bibinfo{pages}{3186--3195}.
\newblock


\bibitem[\protect\citeauthoryear{Zhao, Jia, and Koltun}{Zhao
  et~al\mbox{.}}{2020b}]%
        {esair2020cvpr}
\bibfield{author}{\bibinfo{person}{Hengshuang Zhao}, \bibinfo{person}{Jiaya
  Jia}, {and} \bibinfo{person}{Vladlen Koltun}.}
  \bibinfo{year}{2020}\natexlab{b}.
\newblock \showarticletitle{Exploring self-attention for image recognition}. In
  \bibinfo{booktitle}{\emph{Proceedings of the IEEE/CVF Conference on Computer
  Vision and Pattern Recognition}}. \bibinfo{pages}{10076--10085}.
\newblock


\bibitem[\protect\citeauthoryear{Zhao, Tian, Sun, Shao, Yan, Yi, Wang, and
  Tang}{Zhao et~al\mbox{.}}{2017b}]%
        {spindle2017cvpr}
\bibfield{author}{\bibinfo{person}{Haiyu Zhao}, \bibinfo{person}{Maoqing Tian},
  \bibinfo{person}{Shuyang Sun}, \bibinfo{person}{Jing Shao},
  \bibinfo{person}{Junjie Yan}, \bibinfo{person}{Shuai Yi},
  \bibinfo{person}{Xiaogang Wang}, {and} \bibinfo{person}{Xiaoou Tang}.}
  \bibinfo{year}{2017}\natexlab{b}.
\newblock \showarticletitle{Spindle net: Person re-identification with human
  body region guided feature decomposition and fusion}. In
  \bibinfo{booktitle}{\emph{Proceedings of the IEEE conference on computer
  vision and pattern recognition}}. \bibinfo{pages}{1077--1085}.
\newblock


\bibitem[\protect\citeauthoryear{Zhao, Li, Zhuang, and Wang}{Zhao
  et~al\mbox{.}}{2017a}]%
        {pa2017iccv}
\bibfield{author}{\bibinfo{person}{Liming Zhao}, \bibinfo{person}{Xi Li},
  \bibinfo{person}{Yueting Zhuang}, {and} \bibinfo{person}{Jingdong Wang}.}
  \bibinfo{year}{2017}\natexlab{a}.
\newblock \showarticletitle{Deeply-learned part-aligned representations for
  person re-identification}. In \bibinfo{booktitle}{\emph{Proceedings of the
  IEEE international conference on computer vision}}.
  \bibinfo{pages}{3219--3228}.
\newblock


\bibitem[\protect\citeauthoryear{Zhao, Gao, Zhang, Cheng, Han, Jiang, Guo,
  Zheng, Sang, and Sun}{Zhao et~al\mbox{.}}{2020a}]%
        {dndm2020eccv}
\bibfield{author}{\bibinfo{person}{Shizhen Zhao}, \bibinfo{person}{Changxin
  Gao}, \bibinfo{person}{Jun Zhang}, \bibinfo{person}{Hao Cheng},
  \bibinfo{person}{Chuchu Han}, \bibinfo{person}{Xinyang Jiang},
  \bibinfo{person}{Xiaowei Guo}, \bibinfo{person}{Wei-Shi Zheng},
  \bibinfo{person}{Nong Sang}, {and} \bibinfo{person}{Xing Sun}.}
  \bibinfo{year}{2020}\natexlab{a}.
\newblock \showarticletitle{Do Not Disturb Me: Person Re-identification Under
  the Interference of Other Pedestrians}. In \bibinfo{booktitle}{\emph{European
  Conference on Computer Vision}}. Springer, \bibinfo{pages}{647--663}.
\newblock


\bibitem[\protect\citeauthoryear{Zhao, Yan, Huang, and Li}{Zhao
  et~al\mbox{.}}{2021}]%
        {zhao2021graph}
\bibfield{author}{\bibinfo{person}{Yifan Zhao}, \bibinfo{person}{Ke Yan},
  \bibinfo{person}{Feiyue Huang}, {and} \bibinfo{person}{Jia Li}.}
  \bibinfo{year}{2021}\natexlab{}.
\newblock \showarticletitle{Graph-Based High-Order Relation Discovery for
  Fine-Grained Recognition}. In \bibinfo{booktitle}{\emph{Proceedings of the
  IEEE/CVF Conference on Computer Vision and Pattern Recognition}}.
  \bibinfo{pages}{15079--15088}.
\newblock


\bibitem[\protect\citeauthoryear{Zheng, Shen, Tian, Wang, Wang, and Tian}{Zheng
  et~al\mbox{.}}{2015}]%
        {market2015iccv}
\bibfield{author}{\bibinfo{person}{Liang Zheng}, \bibinfo{person}{Liyue Shen},
  \bibinfo{person}{Lu Tian}, \bibinfo{person}{Shengjin Wang},
  \bibinfo{person}{Jingdong Wang}, {and} \bibinfo{person}{Qi Tian}.}
  \bibinfo{year}{2015}\natexlab{}.
\newblock \showarticletitle{Scalable person re-identification: A benchmark}. In
  \bibinfo{booktitle}{\emph{Proceedings of the IEEE international conference on
  computer vision}}. \bibinfo{pages}{1116--1124}.
\newblock


\bibitem[\protect\citeauthoryear{Zheng, Zheng, and Yang}{Zheng
  et~al\mbox{.}}{2017}]%
        {duke2017iccv}
\bibfield{author}{\bibinfo{person}{Zhedong Zheng}, \bibinfo{person}{Liang
  Zheng}, {and} \bibinfo{person}{Yi Yang}.} \bibinfo{year}{2017}\natexlab{}.
\newblock \showarticletitle{Unlabeled samples generated by gan improve the
  person re-identification baseline in vitro}. In
  \bibinfo{booktitle}{\emph{Proceedings of the IEEE International Conference on
  Computer Vision}}. \bibinfo{pages}{3754--3762}.
\newblock


\bibitem[\protect\citeauthoryear{Zhong, Zheng, Kang, Li, and Yang}{Zhong
  et~al\mbox{.}}{2020}]%
        {re2020aaai}
\bibfield{author}{\bibinfo{person}{Zhun Zhong}, \bibinfo{person}{Liang Zheng},
  \bibinfo{person}{Guoliang Kang}, \bibinfo{person}{Shaozi Li}, {and}
  \bibinfo{person}{Yi Yang}.} \bibinfo{year}{2020}\natexlab{}.
\newblock \showarticletitle{Random erasing data augmentation}. In
  \bibinfo{booktitle}{\emph{Proceedings of the AAAI Conference on Artificial
  Intelligence}}, Vol.~\bibinfo{volume}{34}. \bibinfo{pages}{13001--13008}.
\newblock


\bibitem[\protect\citeauthoryear{Zhou, Yang, Cavallaro, and Xiang}{Zhou
  et~al\mbox{.}}{2019}]%
        {osnet2019iccv}
\bibfield{author}{\bibinfo{person}{Kaiyang Zhou}, \bibinfo{person}{Yongxin
  Yang}, \bibinfo{person}{Andrea Cavallaro}, {and} \bibinfo{person}{Tao
  Xiang}.} \bibinfo{year}{2019}\natexlab{}.
\newblock \showarticletitle{Omni-scale feature learning for person
  re-identification}. In \bibinfo{booktitle}{\emph{Proceedings of the IEEE/CVF
  International Conference on Computer Vision}}. \bibinfo{pages}{3702--3712}.
\newblock


\bibitem[\protect\citeauthoryear{Zhu, Guo, Liu, Tang, and Wang}{Zhu
  et~al\mbox{.}}{2020}]%
        {ig2020eccv}
\bibfield{author}{\bibinfo{person}{Kuan Zhu}, \bibinfo{person}{Haiyun Guo},
  \bibinfo{person}{Zhiwei Liu}, \bibinfo{person}{Ming Tang}, {and}
  \bibinfo{person}{Jinqiao Wang}.} \bibinfo{year}{2020}\natexlab{}.
\newblock \showarticletitle{Identity-Guided Human Semantic Parsing for Person
  Re-Identification}.
\newblock \bibinfo{journal}{\emph{European Conference on Computer Vision}}
  (\bibinfo{year}{2020}).
\newblock


\bibitem[\protect\citeauthoryear{Zhuo, Chen, Lai, and Wang}{Zhuo
  et~al\mbox{.}}{2018}]%
        {occ2018icme}
\bibfield{author}{\bibinfo{person}{Jiaxuan Zhuo}, \bibinfo{person}{Zeyu Chen},
  \bibinfo{person}{Jianhuang Lai}, {and} \bibinfo{person}{Guangcong Wang}.}
  \bibinfo{year}{2018}\natexlab{}.
\newblock \showarticletitle{Occluded person re-identification}. In
  \bibinfo{booktitle}{\emph{2018 IEEE International Conference on Multimedia
  and Expo (ICME)}}. IEEE, \bibinfo{pages}{1--6}.
\newblock


\end{thebibliography}

\end{document}